\PassOptionsToPackage{usenames,dvipsnames,table}{xcolor}

\documentclass[11pt,a4paper]{article}
\usepackage[hyperref]{acl2021}
\usepackage{times}
\usepackage{latexsym}

\usepackage{microtype}

\aclfinalcopy 

\usepackage{url}
\usepackage{times}
\usepackage{latexsym}
\usepackage{url}
\usepackage{amssymb}
\usepackage{array}
\usepackage{amsmath}
\usepackage{algorithm}
\usepackage[noend]{algpseudocode}
\usepackage{graphicx}
\usepackage{url}
\usepackage{times}
\usepackage{helvet}
\usepackage{courier}
\usepackage{balance}  
\usepackage{mdwlist}
\usepackage{graphics}
\usepackage{color, colortbl}
\usepackage{rotating}
\usepackage{epsfig}
\usepackage{alltt}
\usepackage{tabularx}
\usepackage{cellspace}
\usepackage{array}
\usepackage{moreverb}
\usepackage{fancyvrb}
\usepackage{enumerate}
\usepackage{xspace}
\usepackage{relsize}
\usepackage{amsfonts}
\usepackage{txfonts}
\usepackage[labelfont=bf]{caption}
\usepackage{dsfont}
\usepackage{rotating}
\usepackage{latexsym}
\usepackage{multirow}
\usepackage{booktabs}
\usepackage{float}
\usepackage{array,etoolbox}
\usepackage{enumitem}
\usepackage{subfig}
\usepackage{arydshln}
\usepackage{setspace}
\usepackage[page]{appendix}
\usepackage{multirow}
\usepackage{seqsplit}
\usepackage{mathptmx}
\usepackage{anyfontsize}
\usepackage{t1enc}
\usepackage{adjustbox}
\usepackage{chngpage}
\usepackage{amsmath}
\usepackage{bbm}
\usepackage{makecell}
\usepackage{ragged2e}
 \usepackage{afterpage}
 \usepackage{capt-of}
 \definecolor{Gray}{gray}{0.9}
 \definecolor{LightCyan}{rgb}{0.88,1,1}
 \usepackage{ upgreek }
\usepackage{framed}
\usepackage{multicol}

\usepackage{multirow}
\usepackage{extarrows}

\usepackage{stackengine}
\usepackage{rotating}

\newcolumntype{Z}{>{\raggedright\let\newline\\\arraybackslash\hspace{0pt}}X}

\newcolumntype{x}[1]
{>{\raggedright}p{#1}}
\newcolumntype{z}[1]
{>{\centering}p{#1}}

\usepackage{color}
\usepackage[dvipsnames]{xcolor}

\newtoggle{draft}
\toggletrue{draft}
\iftoggle{draft}{
\newcommand{\taehee}[1]{\textcolor{BurntOrange}{[#1 \textsc{--taehee}]}}
\newcommand{\dk}[1]{\textcolor{Maroon}{[#1 \textsc{--dk}]}}
\newcommand{\ed}[1]{\textcolor{Blue}{[#1 \textsc{--ed}]}}
}{
\newcommand{\taehee}[1]{}
\newcommand{\dk}[1]{}
\newcommand{\ed}[1]{}
}

\newcommand{\com}[1]{}

\newcommand{\method}{\textsc{x}\texttt{SLUE}\xspace}
\newcolumntype{P}[1]{>{\centering\arraybackslash}p{#1}}

\title{
Style is NOT a single variable: \\Case Studies for Cross-Style Language Understanding
}

\author{Dongyeop Kang\thanks{$^*$This work was done while DK was at CMU.} \\ \texttt{dongyeopk@berkeley.edu} \\ \texttt{UC Berkeley}
        \And  
        Eduard Hovy \\ \texttt{hovy@cs.cmu.edu} \\ Carnegie Mellon University}
\date{}

\begin{document}
\maketitle
\begin{abstract}
Every natural text is written in some style.  
Style is formed by a complex combination of different stylistic factors, including formality markers, emotions, metaphors, etc. 
One cannot form a complete understanding of a text without considering these factors.  
The factors combine and co-vary in complex ways to form styles. 
Studying the nature of the co-varying combinations sheds light on stylistic language in general, sometimes called \textit{cross-style language understanding}.
This paper provides the benchmark corpus (\method) that combines existing datasets and collects a new one for sentence-level cross-style language understanding and evaluation.
The benchmark contains text in 15 different styles under the proposed four theoretical groupings: figurative, personal, affective, and interpersonal groups.
For valid evaluation, we collect an additional diagnostic set by annotating all 15 styles on the same text.
Using \method, we propose three interesting cross-style applications in classification, correlation, and generation.
First, our proposed cross-style classifier trained with multiple styles together helps improve overall classification performance against individually-trained style classifiers.
Second, our study shows that some styles are highly dependent on each other in human-written text.
Finally, we find that combinations of some contradictive styles likely generate stylistically less appropriate text. 
We believe our benchmark and case studies help explore interesting future directions for cross-style research.
The preprocessed datasets and code are publicly available.\footnote{\url{https://github.com/dykang/xslue}}

\end{abstract}

\section{Introduction}\label{sec:intro}

People often use style as a strategic choice for their personal or social goals in communication \cite{hovy1987generating,silverstein2003indexical,jaffe2009stance,kang2020thesis}.
Some stylistic choices implicitly reflect the author's characteristics, like personality, demographic traits \cite{kang19emnlp_pastel}, and emotions \cite{buechel2017emobank}, whereas others are explicitly controlled by the author's choices for their social goals like using polite language, for better relationship with the elder \cite{danescu2013computational}. 
In this work, we broadly call each individual linguistic phenomena as one specific type of \textit{style}.

Style is not a single variable, but multiple variables have their own degrees of freedom and they co-vary together.
Imagine an orchestra, as a metaphor of style. 
What we hear from the orchestra is the harmonized sound of complex combinations of individual instruments played. 
A conductor, on top of it, controls their combinatory choices among them, such as tempo or score.
Some instruments under the same category, such as violin and cello for bowed string type, make a similar pattern of sound. 
Similarly, text reflects complex combination of multiple styles. Each has its own lexical and syntactic features and some are dependent on each other.
Consistent combination of them by the author will produce stylistically appropriate text.

To the best of our knowledge, only a few recent works have studied style inter-dependencies in a very limited range such across demographical traits \cite{nguyen-etal-2014-gender,preoctiuc2018user}, across emotions \cite{warriner2013norms}, across lexical styles \cite{brooke-hirst-2013-multi}, across genres \cite{passonneau-etal-2014-biber}, or between metaphor and emotion \cite{Dankers2019ModellingTI,mohammad-etal-2016-metaphor}.

Unlike the prior works, this work proposes the first comprehensive understanding of cross-stylistic language variation, particularly focusing on how different styles co-vary together in written text, which styles are dependent on each other, and how they are systematically composed to generate text.

Our work has following contributions:
\begin{itemize}[noitemsep,topsep=0pt,leftmargin=*]
    \item Aggregate 15 different styles and 23 sentence-level classification tasks (\S\ref{sec:cross_style_dataset}). Based on their social goals, the styles are categorized into four groups (Table \ref{tab:category_style}): figurative, affective, personal and interpersonal.
    \item Collect a cross-style set by annotating 15 styles on the same text for valid evaluation of cross-stylistic variation (\S\ref{sec:cross_style_diagnostic_set}). 
    \item Study cross-style variations in classification (\S\ref{sec:style_classification}), correlation (\S\ref{sec:style_correlation}), and generation (\S\ref{sec:cross_style_generation}):
    \begin{itemize}[noitemsep,topsep=0pt,leftmargin=*]
    \item our jointly trained classifier on multiple styles shows better performance than individually-trained classifiers. 
    \item our correlation study finds statistically significant style inter-dependencies  (e.g., impoliteness and offense) in written text. 
    \item our conditional stylistic generator shows that better style classifier enables stylistically better generation. Also, some styles (e.g., impoliteness and positive sentiment) are condtradictive in generation.
    \end{itemize}
\end{itemize}
\section{Related Work}\label{sec:related}

\paragraph{Definition of style.}

People may have different definitions in what they call `style'.
Several sociolinguistic theories on styles have been developed focusing on their inter-personal perspectives, such as Halliday's systemic functional linguistics \cite{halliday2006linguistic} or Biber's theory on register, genre, and style \cite{biber2019register}.

In sociolinguistics, indexicality \cite{silverstein2003indexical,coupland2007style,johnstone2010locating} is the phenomenon where a sign points to some object, but only in the context in which it occurs. 
Nonreferential indexicalities include the speaker's gender, affect \cite{besnier1990language}, power, solidarity \cite{brown1960pronouns}, social class, and identity \cite{ochs1990indexicality}.

Building on Silverstein's notion of indexical order, \citet{eckert2008variation} built the notion that linguistic variables index a social group, which leads to the indexing of certain traits stereotypically associated with members of that group. 
\citet{Eckert2000LinguisticVA,eckert2019limits} argued that style change creates a new persona, impacting a social landscape and presented the expression of social meaning as a continuum of decreasing reference and increasing performativity. 


Despite the extensive theories, very little is known on extra-dependency across multiple styles.
In this work, we empirically show evidence of extra-linguistic variations of styles, like a formality, politeness, etc, but limited to styles only if we can obtain \textit{publicly available resources for computing}.
We call the individual phenomena a specific type of ``style'' in this work. 
We admit that there are many other kinds of styles not covered in this work, such as inter-linguistic variables in grammars and phonology, or high-level style variations like individual's writing style or genres.

\paragraph{Cross-style analysis.}
Some recent works have provided empirical evidence of style inter-dependencies but in a very limited range: 
\citet{warriner2013norms} analyzed emotional norms and their correlation in lexical features of text.
\citet{chhaya-etal-2018-frustrated} studied a correlation of formality, frustration, and politeness but on small samples (i.e., 960 emails).
\citet{nguyen-etal-2014-gender} focused on correlation across demographic information (e.g., gender, age) and with some other factors such as emotions \cite{preoctiuc2018user}.
\citet{Dankers2019ModellingTI,mohammad-etal-2016-metaphor} studied the interplay of metaphor and emotion in text.
\citet{liu2010sentiment} studied sarcasm detection using sentiment as a sub-problem. 
\citet{brooke-hirst-2013-multi} conducted a topical analysis of six styles: literary, abstract, objective, colloquial, concrete, and subjective, on different genres of text.
\citet{passonneau-etal-2014-biber} conducted a detailed analysis of Biber's  genres and relationship between genres.

\section{\method: A Benchmark for Cross-Style Language Understanding and Evaluation
}\label{sec:cross_style_dataset}



\begin{table}[t]
\centering
\small
\begin{tabularx}{\columnwidth}{@{}p{2.0cm}p{5.2cm}@{}}
\toprule
\textbf{Groups}  & \textbf{Styles}\\
\midrule
\textsc{Interpersonal}  &
{{Formality}, {Politeness}}\\
\hdashline[0.4pt/2pt]
\textsc{Figurative} &
{{Humor}, {Sarcasm}, {Metaphor}}\\
\hdashline[0.4pt/2pt]
\textsc{Affective} &
{{Emotion}, {Offense}, {Romance}, {Sentiment}}\\
\hdashline[0.4pt/2pt]
\textsc{Personal} &
{{Age}, {Ethnicity}, {Gender}, {Education level}, {Country}, {Political view}}\\
\bottomrule
\end{tabularx}
\caption{\label{tab:category_style} Style grouping in \method. } 
\end{table}

\begin{table*}[th]
\centering
\resizebox{\linewidth}{!}{%
\begin{tabular}{@{}c lcccp{6.5cm}ccc>{\columncolor{white}[\tabcolsep][0pt]}c@{}}
\toprule
&
\textbf{\textbf{Style} \& dataset} & \textbf{\#S} & \textbf{Split} & \textbf{\#L} & \textbf{Label (distribution)} & \textbf{B} & \textbf{Domain} & \textbf{Public} & \textbf{Task} \\
\toprule
\addlinespace[0.1cm]
\parbox[t]{2mm}{\multirow{4}{*}{\rotatebox[origin=c]{90}{\textsc{Interpers.}}}} 
& \textbf{Formality} \\
&
\texttt{GYAFC} \cite{rao2018dear} & 224k & given & 2 & formal (50\%), informal (50\%) & Y & web & N & clsf. \\
\cmidrule(lr{0em}){2-10}
& \textbf{Politeness} \\
& \texttt{StanfPolite} \cite{danescu2013computational} & 10k & given & 2 & polite (49.6\%), impolite (50.3\%) & Y & web & Y & clsf.\\
\midrule
\parbox[t]{2mm}{\multirow{10}{*}{\rotatebox[origin=c]{90}{\textsc{Figurative}}}} 
& \textbf{Humor}\\
& \texttt{ShortHumor} \cite{shorthumor} & 44k & random & 2 & humor (50\%), non-humor (50\%) & Y & web & Y & clsf.  \\
& \texttt{ShortJoke} \cite{shortjoke} & 463k & random & 2 & humor (50\%), non-humor (50\%) & Y & web & Y & clsf. \\
\cmidrule(lr{0em}){2-10}
& \textbf{Sarcasm}\\
& \texttt{SarcGhosh} \cite{ghosh2016fracking} & 43k & given & 2 & sarcastic (45\%), non-sarcastic (55\%) & Y & tweet  & Y & clsf. \\
& \texttt{SARC} \cite{khodak2017large} & 321k & given & 2 & sarcastic (50\%), non-sarcastic (50\%) & Y & reddit & Y & clsf. \\
& \texttt{SARC}\_pol \cite{khodak2017large} & 17k & given & 2 & sarcastic (50\%), non-sarcastic (50\%) & Y & reddit & Y & clsf. \\
\cmidrule(lr{0em}){2-10}
& \textbf{Metaphor}\\
& \texttt{VUA} \cite{steen2010method}  & 23k & given & 2 & metaphor (28.3\%), non-metaphor (71.6\%) & N & misc. & Y & clsf. \\
& \texttt{TroFi} \cite{birke2006clustering}  & 3k & random & 2 &  metaphor (43.5\%), non-metaphor (54.5\%) & N & news & Y & clsf. \\
\midrule
\parbox[t]{2mm}{\multirow{12}{*}{\rotatebox[origin=c]{90}{\textsc{Affective}}}} &
\textbf{Emotion}  \\
& \texttt{EmoBank}$_{valence}$ \cite{buechel2017emobank} & 10k & random & 1 & negative, positive & - & misc. & Y & rgrs.\\
& \texttt{EmoBank}$_{arousal}$ \cite{buechel2017emobank} & 10k & random & 1 & calm, excited & - & misc. & Y & rgrs.\\
& \texttt{EmoBank}$_{dominance}$ \cite{buechel2017emobank} & 10k & random & 1 & being\_controlled, being\_in\_control & - & misc. & Y & rgrs.\\
& \texttt{DailyDialog} \cite{li2017dailydialog} & 102k & given & 7 & noemotion(83\%), happy(12\%).. & N & dialogue & Y & clsf.\\
\cmidrule(lr{0em}){2-10}
& \textbf{Offense} \\
& \texttt{HateOffensive} \cite{davidson2017automated} & 24k & given & 3 & hate(6.8\%), offensive(76.3\%).. & N & tweet & Y & clsf. \\ 
\cmidrule(lr{0em}){2-10}
& \textbf{Romance} \\
& \texttt{ShortRomance} & 2k & random  & 2 & romantic (50\%), non-romantic (50\%) & Y & web & Y & clsf.\\
\cmidrule(lr{0em}){2-10}
& \textbf{Sentiment} \\
& \texttt{SentiBank} \cite{socher2013recursive} & 239k & given  & 2 & positive (54.6\%), negative (45.4\%) & Y & web & Y & clsf.\\
\midrule
\parbox[t]{2mm}{\multirow{6}{*}{\rotatebox[origin=c]{90}{\textsc{Personal}}}} &
\textbf{Gender} \texttt{PASTEL} \cite{kang19emnlp_pastel} & 41k & given & 3 & Female (61.2\%), Male (38.0\%).. & N & caption & Y & clsf.\\
&\textbf{Age} \texttt{PASTEL} \cite{kang19emnlp_pastel} & 41k & given  & 8 & 35-44 (15.3\%), 25-34 (42.1\%).. & N & caption & Y & clsf.\\ 
&\textbf{Country} \texttt{PASTEL} \cite{kang19emnlp_pastel} & 41k & given & 2 & USA (97.9\%), UK (2.1\%) & N & caption & Y & clsf.\\
&\textbf{Politics} \texttt{PASTEL} \cite{kang19emnlp_pastel} & 41k & given & 3 & LeftWing (42.7\%), Centerist(41.7\%).. & N & caption & Y & clsf.\\ 
&\textbf{Education} \texttt{PASTEL} \cite{kang19emnlp_pastel} & 41k & given & 10 & Bachelor(30.6\%), Master(18.4\%)..	& N & caption & Y & clsf.\\ 
&\textbf{Ethnicity} \texttt{PASTEL} \cite{kang19emnlp_pastel} & 41k & given & 10 & Caucasian(75.6\%), African(5.5\%).. & N & caption & Y & clsf. \\ 
\bottomrule
\end{tabular}\vspace{-4mm}
}
\caption{\label{tab:datasets} Style datasets in \method. 
\#S and \#L mean the number of total samples and labels, respectively. 
\textbf{B} means whether the labels are balanced (Y) or not (N).
Every label is normalized, rangin   g in $[0,1]$. 
\textbf{Public} means whether dataset is publicly available or not.
clsf. and rgrs. in Task denotes classification and regression, respectively. 
} 
\end{table*}

\subsection{Style selection and groupings}\label{sec:style_selection_grouping}
In order to conduct a comprehensive style research, one needs to collect a collection of different style datasets.
We survey recent papers related to style research published in ACL venues and choose 15 widely-used styles that have publicly available annotated resources and feasible size of training dataset (Table \ref{tab:category_style}).
We plan to gradually increase the coverage of style kinds and make the benchmark more comprehensive in the future.

We follow the theoretical style grouping criteria based on their social goals in \citet{kang2020thesis} that categorizes styles into four groups (Table \ref{tab:category_style}): \textsc{personal}, \textsc{interpersonal}, \textsc{figurative}, and \textsc{affective} group, where each group has its own social goals in communication.
This grouping will be used in our case studies as a basic framework to detect their dependencies.


\subsection{Individual style dataset}
For each style in the group, we pre-process existing style datasets or collect our own if there is no publicly available one (i.e., \texttt{ShortRomance}).
We do not include datasets with small samples (e.g., $\leq1$K) due to its infeasibility of training a large model.
We also limit our dataset to classify a single sentence, although there exists other types of datasets (e.g., document-level style classifications, classifying a sentence with respect to context given) which are out of scope of this work.

If a dataset has its own data split, we follow that. Otherwise, we randomly split it by 0.9/0.05/0.05 ratios for the train, valid, and test set, respectively.
If a dataset has only positive samples (\texttt{ShortHumor}, \texttt{ShortJoke}, \texttt{ShortRomance}), we do negative sampling from literal text as in \citet{khodak2017large}.
We include the detailed pre-processing steps in Appendix \S\ref{sec:style_dataset}.

\subsection{Cross-style diagnostic set}\label{sec:cross_style_diagnostic_set}
The individual datasets, however, have variations in domains (e.g., web, dialogue, tweets), label distributions, and data sizes (See domain, label, and \#S columns in Table \ref{tab:datasets}).
Evaluating a system with these individual datasets' test set is not an appropriate way to validate how multiple styles are used together in a mixed way, because samples from individual datasets are annotated only when a single style is considered.

To help researchers evaluate their systems in the cross-style setting, we collect an additional diagnostic set, called \textit{cross-set} by annotating labels of 15 styles together on the same text from crowd workers. 
We collect total 500 sample texts from two different sources:
the first half is randomly chosen from test sets among the 15 style datasets in balance, and the second half is chosen from random tweets that have high variations across style prediction scores using our pre-trained style classifiers.
Each sample text is annotated by five annotators, and the final label for each style is decided via majority voting over the five annotations.
In case they are tied or all different from each other for multiple labels, we don't include them.
We also include \underline{Don't Know} option for personal styles and \underline{Neutral} option for two opposing binary styles (e.g., sentiment, formality).
The detailed annotation schemes are in Appendix \S\ref{sec:annotation_scheme}.

\begin{table}[t!]
\centering
\small
\begin{tabular}{rlrl}
\cellcolor{gray!15}Sentiment	&	\cellcolor{gray!15}0.81	&	\cellcolor{gray!60}Sarcasm	&	\cellcolor{gray!60}0.38\\
\cellcolor{gray!15}Politeness	&	\cellcolor{gray!15}0.75	&	\cellcolor{gray!60}Country	&	\cellcolor{gray!60}0.38\\
\cellcolor{gray!35}Formality	&	\cellcolor{gray!35}0.48	&	\cellcolor{gray!60}Humor	&	\cellcolor{gray!60}0.37\\
\cellcolor{gray!35}Gender	&	\cellcolor{gray!35}0.47	&	\cellcolor{gray!60}Education level	&	\cellcolor{gray!60}0.36\\
\cellcolor{gray!35}Emotion: Valence	&	\cellcolor{gray!35}0.43	&	\cellcolor{gray!60}Age	&	\cellcolor{gray!60}0.35\\
\cellcolor{gray!35}Emotion	&	\cellcolor{gray!35}0.42	&	\cellcolor{gray!60}Political view	&	\cellcolor{gray!60}0.32\\
\cellcolor{gray!35}Romance	&	\cellcolor{gray!35}0.42	&	\cellcolor{gray!60}Metaphor	&	\cellcolor{gray!60}0.29\\
\cellcolor{gray!35}Offense	&	\cellcolor{gray!35}0.41	&	\cellcolor{gray!60}Emotion: Arousal	&	\cellcolor{gray!60}0.26\\
\cellcolor{gray!35}Ethnicity	&	\cellcolor{gray!35}0.41	&	\cellcolor{gray!60}Emotion: Dominance	&	\cellcolor{gray!60}0.24\\
\end{tabular}
\caption{\label{tab:style_annotation_agreement}
Annotator's agreement (Krippendorff's alpha).
The degree of gray shading shows {\colorbox{gray!15}{good}}, {\colorbox{gray!35}{moderate}}, and {\colorbox{gray!60}{fair}} agreements.} 
\end{table}

\begin{table*}[th!]
\centering
\small
\begin{tabular}{@{}
c  @{\hskip 0.2cm}l @{\hskip -0.5cm}r 
c@{\hskip 0.2cm}c@{\hskip 0.2cm}c@{\hskip 0.2cm}c@{\hskip 0.2cm}c@{\hskip 0.4cm}c
c@{\hskip 0.2cm}c@{\hskip 0.4cm}c
@{}}
\toprule
&&\textit{Evaluation set} $\rightarrow$ &  \multicolumn{ 6}{c}{\textit{Individual-set evaluation}} & \multicolumn{ 3}{c}{\textit{Cross-set evaluation} (\S\ref{sec:cross_style_diagnostic_set})} \\
\cmidrule(rr){2-3} \cmidrule(rr){4-9} \cmidrule(lr){10-12}
&&\texttt{Models} $\rightarrow$ &  \multicolumn{ 5}{c}{\texttt{single}} & \texttt{cross} & \multicolumn{ 2}{c}{\texttt{single}} & \texttt{cross} \\
\cmidrule(rr){2-3} \cmidrule(rr){4-8} \cmidrule(rr){9-9} \cmidrule(lr){10-11} \cmidrule(lr){12-12}
& Style $\downarrow$ & \texttt{Dataset} $\downarrow$ &  
\textbf{Majority}&  \textbf{biLSTM} & \textbf{BERT}  & \textbf{RoBERTa} & \textbf{T5} & \textbf{Ours} &
\textbf{BERT} & \textbf{T5} & \textbf{Ours} \\
\midrule
\parbox[t]{2mm}{\multirow{2}{*}{\rotatebox[origin=c]{90}{\textsc{Inter.}}}} 
&
\multirow{1}{*}{\rotatebox[origin=c]{0}{\small{{Formality}}}}&
\texttt{GYAFC}          & 30.2         & 76.4  & 89.4&	89.3&	89.4&	\textbf{89.9}& \textbf{37.3} &	33.8 &	35.0\\
\cmidrule(rr){2-3} \cmidrule(rr){4-9} \cmidrule(lr){10-12}
&\multirow{1}{*}{\rotatebox[origin=c]{0}{\small{{Politeness}}}}&
\texttt{SPolite}    & 36.2  &  61.8 &  68.9	&70.4	&\textbf{71.6}&	71.2 & 60.0 &	62.1 &	\textbf{64.4}\\
\midrule
\parbox[t]{2mm}{\multirow{7}{*}{\rotatebox[origin=c]{90}{\textsc{Figurative}}}} 
&
\multirow{2}{*}{\rotatebox[origin=c]{0}{\small{{Humor}}}}&
\texttt{ShortHumor}     & 33.3        & 88.6 &  97.3&	97.5	&97.4	&\textbf{98.9}& - & - & -\\
&&\texttt{ShortJoke}	    & 33.3    & 89.1 & 98.4	&98.2&	98.5&	\textbf{98.6} & \textbf{50.5} & 	47.2 & 	47.9\\
\cmidrule(rr){2-3} \cmidrule(rr){4-9} \cmidrule(lr){10-12}
&
\multirow{2}{*}{\rotatebox[origin=c]{0}{\small{{Sarcasm}}}}
&
\texttt{SARC}           & 33.3   & 63.0 & 71.5 & 	\textbf{73.1} & 	72.4 & 	72.8 &  \textbf{41.4} & 	37.7 & 	37.4\\
&&\texttt{SARC}{\_pol}   & 33.3      &    61.3 & 73.1 &	\textbf{74.5} &	73.7 &	\textbf{74.4} & - & - & -\\
\cmidrule(rr){2-3} \cmidrule(rr){4-9} \cmidrule(lr){10-12}
&
\multirow{2}{*}{\rotatebox[origin=c]{0}{\small{{Metaphor}}}}&
\texttt{VUA}            & 41.1      & 68.9 & 78.6 &	\textbf{81.4} &	78.9 &	78.0 & \textbf{49.8}	 & 49.0 &	49.1 \\
&&
\texttt{TroFi}          & 36.4        &  73.9 &  \textbf{77.1} &	74.8	 & 76.7 &	76.2  & - & - & -\\
\midrule
\parbox[t]{2mm}{\multirow{8}{*}{\rotatebox[origin=c]{90}{\textsc{Affective}}}} 
&
\multirow{4}{*}{\rotatebox[origin=c]{0}{\small{{Emotion}}}}&
\texttt{EmoBank}$_{Valence}$        & 32.4       & 78.5  & 81.2	 & \textbf{82.8} &	80.8 &	82.5 &- & - & -\\
&&\texttt{EmoBank}$_{Arousal}$        & 34.2       & 49.4  & 58.7 &	\textbf{62.3} &	58.2 &	61.5 &- & - & -\\
&&\texttt{EmoBank}$_{Domin.}$        & 31.3       & 39.5  &  43.6 &	\textbf{48.3} &	42.9 &	46.4 &- & - & - \\
&&\texttt{DailyDialog}    & 12.8    &   27.6 & 48.7 &	46.9 &	\textbf{49.2} &	49.0 & 22.4	& 26.9	& \textbf{33.3}	\\
\cmidrule(rr){2-3} \cmidrule(rr){4-9} \cmidrule(lr){10-12}
&\multirow{1}{*}{\rotatebox[origin=c]{0}{\small{{Offense}}}}&
\texttt{HateOffens}     & 28.5      & 68.2 & 91.9 &	92.4 &	91.7 &	\textbf{93.4} & 34.4 &	36.9 &	\textbf{45.9}\\
\cmidrule(rr){2-3} \cmidrule(rr){4-9} \cmidrule(lr){10-12}
&\multirow{1}{*}{\rotatebox[origin=c]{0}{\small{{Romance}}}}&
\texttt{ShortRomance}   & 33.3        & 90.6  & 99.0 & 	\textbf{100.0} & 98.0 & 	99.0 & \textbf{53.9} &	55.2 &	48.2\\
\cmidrule(rr){2-3} \cmidrule(rr){4-9} \cmidrule(lr){10-12}
&\multirow{1}{*}{\rotatebox[origin=c]{0}{\small{{Sentiment}}}}&
\texttt{SentiBank}  & 33.3       & 82.8 &  96.9 &	\textbf{97.4} &	97.0 &	96.6 & 80.4 &	79.7 &	\textbf{84.6}\\
\midrule
\parbox[t]{2mm}{\multirow{6}{*}{\rotatebox[origin=c]{90}{\textsc{Personal}}}} 
&
Gender&
\texttt{PASTEL}         & 25.7      & 45.5 & 47.7	& 47.9 &	47.3 &	\textbf{50.5} & 29.2 &	32.4 &	\textbf{42.3}  \\
&Age&\texttt{PASTEL}	        & 7.3      &15.2 &  23.0 &	21.7 &	21.3 &	\textbf{23.3} & \textbf{36.1} &	27.0 &	28.1	\\
&Country&\texttt{PASTEL}        & 49.2       & 49.3  & 54.5 &	49.3 &	51.8 &	\textbf{58.4} &	\textbf{49.4} &	46.7 &	48.7 \\
&Political view&\texttt{PASTEL}       & 20.0       &  33.5 & 46.1 &	44.6 &	44.3 &	\textbf{46.7} & \textbf{27.7} &	20.6 &	21.3\\
&Education&\texttt{PASTEL}       & 4.7       & 15.0  &  24.6 &	22.4 &	21.4 &	\textbf{27.3} & 10.3 &	11.4 &	\textbf{15.7}\\
&Ethnicity&\texttt{PASTEL}        & 8.5       &17.6	 &  \textbf{24.4} &	22.5 &	22.4 &	23.8 & \textbf{10.8}	& 8.8 &	9.1\\
\midrule
&&\textbf{Avearge}&26.8 & 56.9	& 64.8&		64.9&		64.2&		\textbf{65.9}& 39.6	 & 38.4 &	\textbf{40.7}\\
\bottomrule
\end{tabular}
\caption{\label{tab:single_style_classify} 
Individual style (left) and cross style (right) classification in \method.
Every score is averaged over ten runs of experiments with different random seeds.
For cross-style classification, we choose a single dataset per style, which has larger training data than the others. Otherwise, we leave it as a blank (-).} 
\end{table*}

Table \ref{tab:style_annotation_agreement} shows annotator's agreement on the cross-set.
We find that annotator's agreement varies a lot depending on style:
sentiment and politeness with good agreement, and formality, emotion, and romance with moderate agreement.
However, personal styles (e.g., age, education level, and political view), metaphor, and emotions (e.g., arousal and dominance), show fair agreements, indicating how difficult and subjective styles they are.

\subsection{Contribution}
Most datasets in \method except for \texttt{Romance} are collected from others' work.
Following the data statement \cite{bender-friedman-2018-data}, we cite and introduce individual datasets with their data statistics in Table \ref{tab:datasets}. 
Our main contribution is to make every dataset to have the same pre-processed format, and distribute them with accompanying code for better reproducibility and accessibility.
Besides this engineering effort, \method's main goal is to invite NLP researchers to the field of cross-style understanding and provide them a valid set of evaluation for further exploration. 
As the first step, using \method, we study cross-style language variation in various applications such as classification (\S\ref{sec:style_classification}), correlation (\S\ref{sec:style_correlation}), and generation (\S\ref{sec:cross_style_generation}).

\section{Case \#1: Cross-Style Classification}\label{sec:style_classification}

We study how modeling multiple styles together, instead of modeling them individually, can be effective in style classification task. 
Particularly, the annotated cross-set in \method will be used as a part of evaluation for cross-style classification.

\paragraph{Models.}

We compare two types of models: \texttt{single} and \texttt{cross} model.
The single model is trained on individual style dataset separately, whereas the cross model is trained on shuffled set of every dataset together. 
For single model, we use various baseline models, such as majority classifier by choosing the majority label in training data, Bidirectional LSTM (biLSTM) \cite{hochreiter1997long} with GloVe embeddings \cite{pennington2014glove}, and variants of fine-tuned transformers; Bidirectional Encoder Representations from Transformers (BERT) \cite{devlin2018bert}, robustly optimized BERT (RoBERTa) \cite{liu2019roberta}, and text-to-text transformer (T5) \cite{raffel2019exploring}.\footnote{For a fair comparison, we restrict size of the pre-trained transformer models to `base` model only, although additional improvement from the larger models is possible.}  
\begin{figure}[t!]
\centering
{
\includegraphics[trim=0cm 9.1cm 15.4cm 0cm,clip,width=0.99\linewidth]{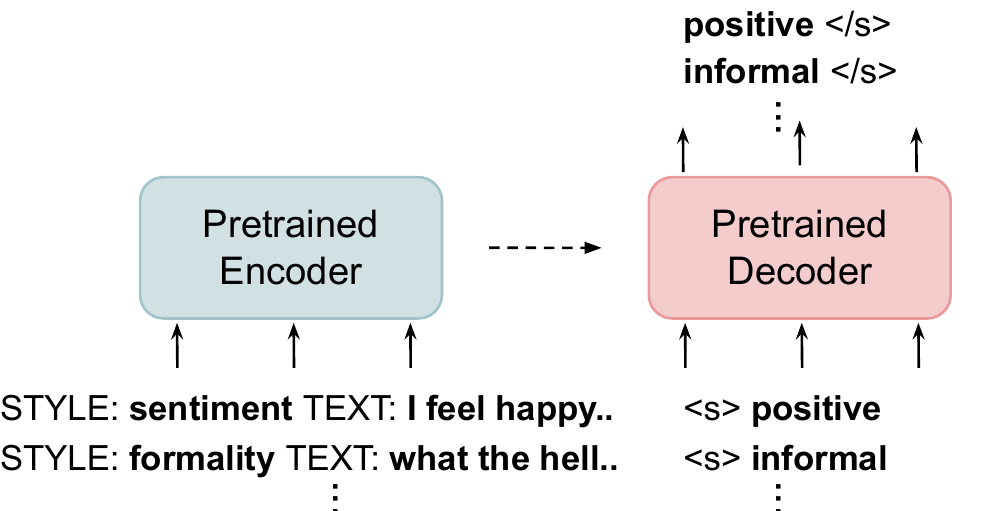}
\vspace{-6mm}
}
\caption{\label{fig:t5x_architecture} Our proposed cross-style classification model. The encoder and decoder are fine-tuned on the combined training datasets in \method.
}
\vspace{-4mm}
\end{figure}

For cross model, we propose an encoder-decoder based model that learns  cross-style patterns with the shared internal representation across styles (Figure \ref{fig:t5x_architecture}). 
It encodes different styles of input as text (e.g., ``STYLE: formality TEXT: would you please..'') and decodes output label as text (e.g., ``formal''). 
We use the pretrained encoder-decoder model from T5 \cite{raffel2019exploring}, and finetune it using the combined, shuffled datasets in \method.
Due to the nature of encoder-decoder model, we can take any training instances for classification tasks into the same text-to-text format.
We also trained the single model (e.g., RoBERTa) on the combined datasets via a multi-task setup (i.e., 15 different heads), but showing less significant result.
The detailed hyper-parameters used in our model training are in Appendix \S\ref{sec:hyper_parameters}.

\paragraph{Tasks.}
Our evaluation has two tasks: \textit{individual-set evaluation} for evaluating a classifier on individual dataset's test set (left columns in Table \ref{tab:single_style_classify}) and \textit{cross-set evaluation} for evaluating a classifier on the annotated cross-set collected in \S\ref{sec:cross_style_diagnostic_set} (right columns in Table \ref{tab:single_style_classify}). 

Due to the label imbalance of datasets, we measure f-score (F1) for classification tasks and Pearson-Spearman correlation for regression tasks (i.e., \texttt{EmoBank}).
For multi-labels, all scores are macro-averaged on each label.

\paragraph{Results.}

In the individual-set evaluation, compared to the biLSTM classifier, the fine-tuned transformers show significant improvements (+8\% points F1) on average, although the different transformer models have similar F1 scores.
Our proposed cross model, significantly outperforms the single model, by +1.7 percentage points overall F1 score, showing the benefit of learning multiple styles together.
Particularly, the cross model significantly improves F1 scores on personal styles such as gender, age, and education level, possibly because the personal styles may be beneficial from detecting other styles.
Among the styles, all personal styles, figurative styles (e.g., sarcasm and metaphor), and emotions are the most difficult styles to predict, which is similarly observed in the annotator's agreement in Table \ref{tab:style_annotation_agreement}. 

In cross-set evaluation, the overall performance significantly drops against the individual set evaluation, like from 65.9\% to 40.7\%, showing why it is important to have these annotated diagnostic set for valid evaluation of cross-style variation.
Again, the cross-style model achieves +1.2\% gain than the single models.
\begin{figure}[th]
\centering
{
\includegraphics[trim=1cm 1cm 1cm 1cm,clip,width=0.8\linewidth]{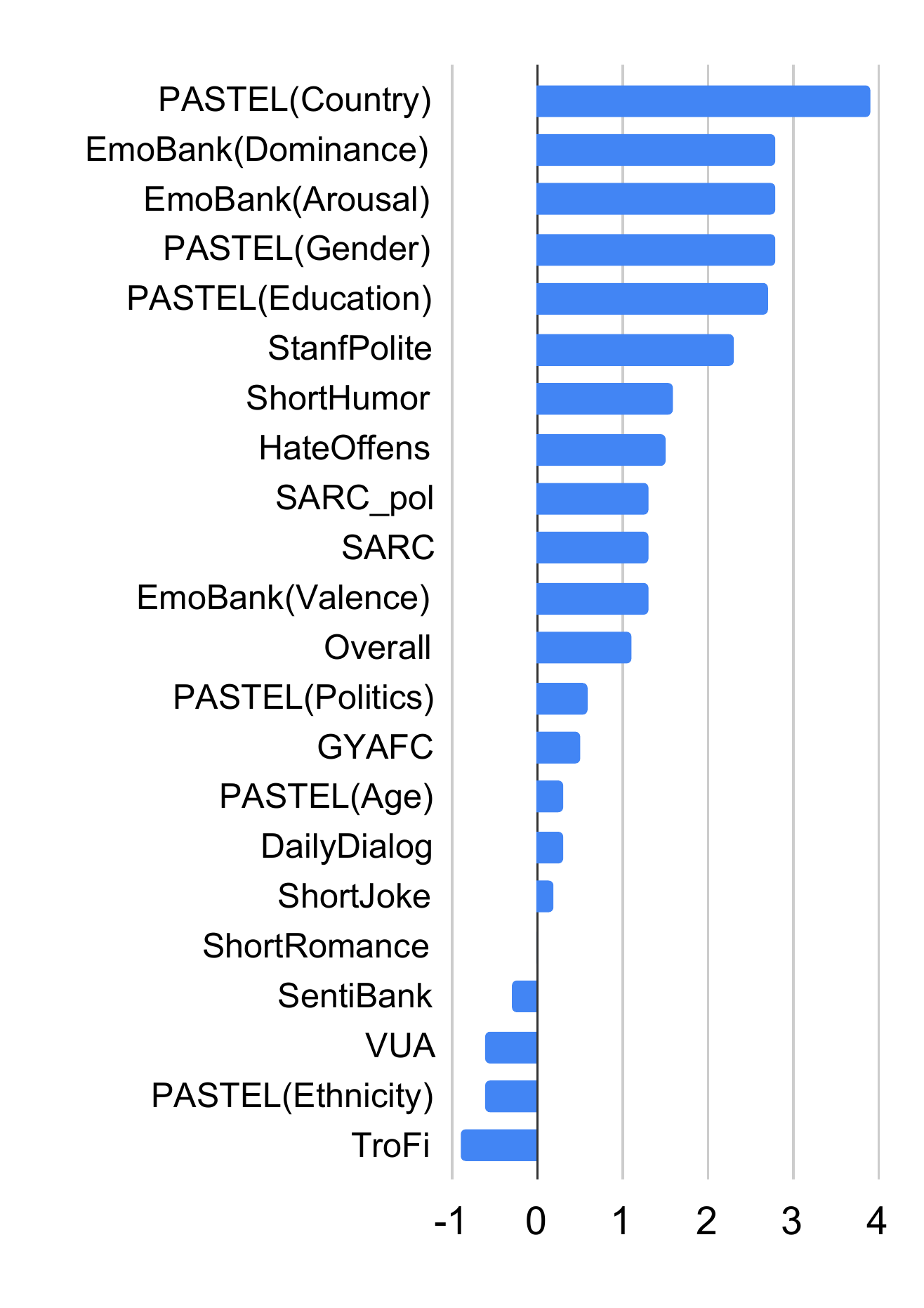}
\vspace{-2mm}
}
\caption{\label{fig:style_classification_diff_individual} F1 improvement by our cross model over BERT in individual style classification task.
\vspace{-2mm}
}
\end{figure}

Figure \ref{fig:style_classification_diff_individual} shows F1 improvement by the cross model against the single model BERT. 
Most styles obtain performance gain from the cross-style modeling, whereas not in the two metaphor style datasets (VUA, TroFi) and ethnicity style. 
This is possibly because metaphor tasks prepend the target metaphor verb to the input text, which is different from other task setups. Thus, learning them together may harm the performance, although it is not significant. 

\section{Case \#2: Style Dependencies}\label{sec:style_correlation}

In addition to the theoretical style grouping in \S\ref{sec:style_selection_grouping}, we empirically find how two styles are correlated in human-written text using silver predictions from the classifiers. 


\paragraph{Setup.}
We sample 1,000,000 tweets crawled using Twitter's Gardenhose API.
We choose tweets as the target domain, because of their stylistic diversity compared to other domains, such as news articles.
Using the fine-tuned cross-style classifier in \S\ref{sec:style_classification}, we predict probability of 53 style attributes\footnote{Attribute means labels of each style: \textit{positive} and \textit{negative} labels for sentiment style.} over the 1M tweets.
We split a tweet into sentences and then average their prediction scores.
We then produce a correlation matrix across the style attributes using Pearson correlation coefficients with Euclidean distance and finally output a $53 \times 53$ correlation matrix.
We only show correlations that are statistical significant with p-value < 0.05 and cross out the rest.

\paragraph{Reliability.}
One may doubt about the classifier's low performance on some styles, leading to unreliable interpretation of our analysis. 
Although we only show correlation on the predicted style values, we also performed the same analysis on the human-annotated cross-set, showing similar correlation tendencies to the predicted ones.
However, due to the small number of annotations, its statistical significance is not high enough.
Instead, we decide to show the prediction-based correlation, possibly including noisy correlations but with statistical significance.


\begin{figure*}[ht!]
\vspace{0mm}
\centering
{

\includegraphics[trim=0 0 0 0,clip,width=0.9\linewidth]{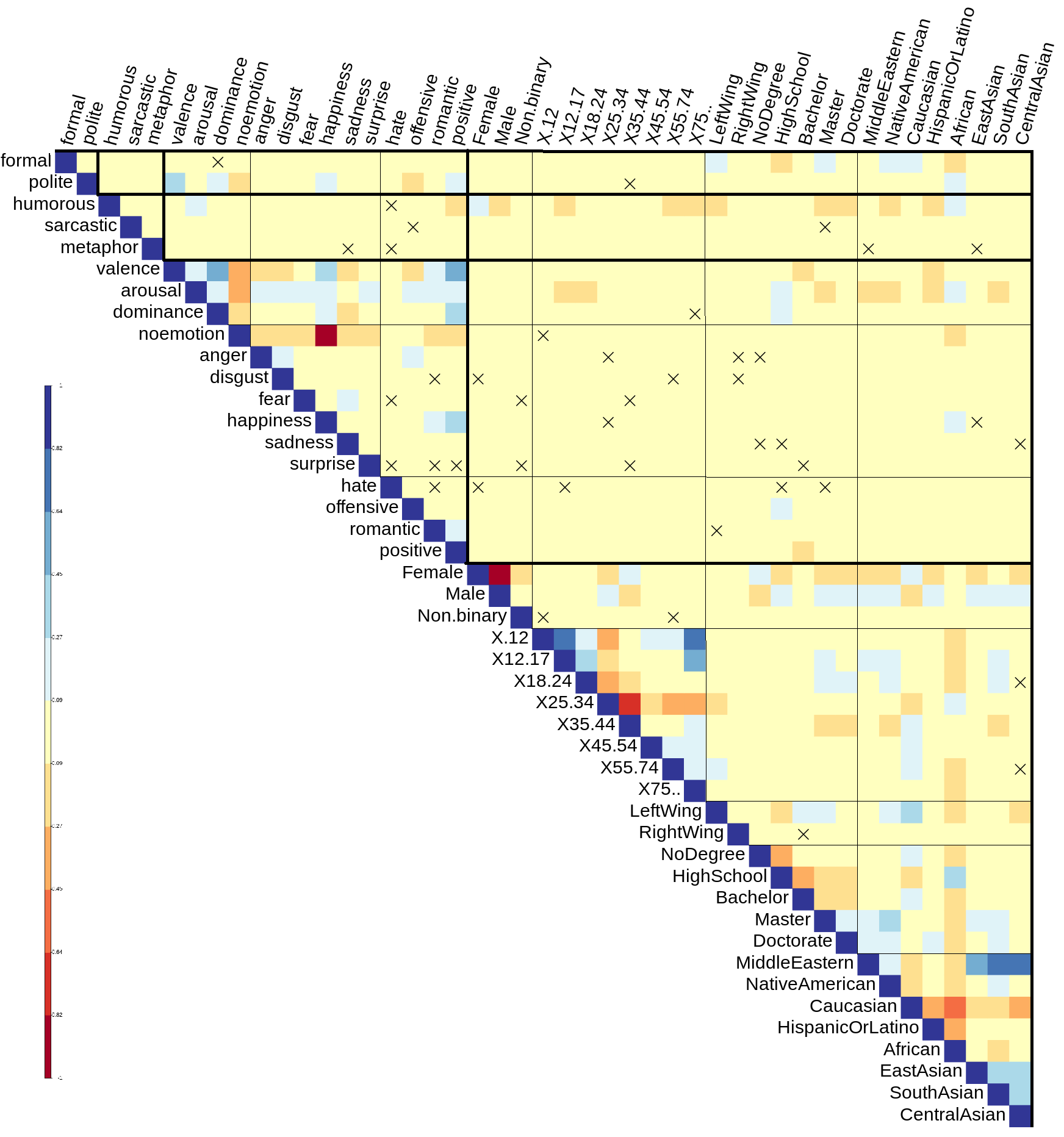}
\vspace{-2mm}
}
\caption{\label{fig:style_corr} Cross-style correlation. 
Correlations with $p < 0.05$ (confidence interval: 0.95) are only considered as statistically significant. 
The degree of correlation gradually increases from red (negative) to blue (positive), where the color intensity is proportional to the correlation coefficients.  
We partition the correlation matrix into three pieces: across interpersonal, figurative and affective styles (upper left), between persona and a group of interpersonal, figurative, and affective styles (upper right), and across persona styles (lower right).
IMPORTANT NOTE: \textit{please be VERY CAREFUL not to make any unethical or misleading interpretations from these model-predicted artificial correlations.}
Best viewed in color.  
\vspace{-4mm}
}
\end{figure*}

\begin{table}[t!]
\centering
\small
\begin{tabular}{@{}p{2.4cm}p{3.3cm}p{0.5cm}@{}}
\toprule
\textbf{Target Style} & \textbf{Correlated styles} & \textbf{H}\\
\midrule
\multirow{2}{*}{Humorous} & Excitement emotion & 5.0\\
&  Negative sentiment & 3.5\\
\hdashline[0.4pt/2pt]
\multirow{2}{*}{Polite} & Positive valence emotion & 4.5\\
&  Happy emotion & 4.0 \\
& \underline{No} offense & 5.0 \\
\hdashline[0.4pt/2pt]
\multirow{2}{*}{Positive sentiment} & Happy emotion & 4.5\\
&  \underline{No} offense & 4.7 \\
& \underline{No} hate & 4.7 \\
\hdashline[0.4pt/2pt]
\multirow{2}{*}{Dominance emotion} & Happy emotion & 3.7\\
& Positive sentiment & 3.7\\
\hdashline[0.4pt/2pt]
\multirow{2}{*}{Anger emotion} & Disgust emotion& 4.0\\
& Offense & 5.0 \\
\hdashline[0.4pt/2pt]
\multirow{2}{*}{Happy emotion} & Romance & 4.7\\
& Positive sentiment & 4.7\\
\hdashline[0.4pt/2pt]
Formal & Master education & 4.0 \\
\hdashline[0.4pt/2pt]
Informal & High-school education & 4.0\\
\hdashline[0.4pt/2pt]
\multirow{2}{*}{Non-humorous} & Age 55< & 3.7\\
& Doctorate education & 4.0\\
\hdashline[0.4pt/2pt]
High-school educ. & Excitement emotion & 2.7\\
& Offense  & 3.0\\
\hdashline[0.4pt/2pt]
Master education & Doctorate education & 4.2 \\
\hdashline[0.4pt/2pt]
Caucasian & \underline{No} Hispanic & 4.2\\
\bottomrule
\end{tabular}
\caption{\label{tab:style_correlation_matrix} Some example pairs of positively (or negatively for ``\underline{No}'') correlated styles with human judgement score (H). } 
\end{table}

\paragraph{Results.}
Figure \ref{fig:style_corr} shows the full correlation matrix we found.
From the matrix, we summarize some of the highly correlated style pairs in Table \ref{tab:style_correlation_matrix} 
For each pair of correlation, two annotators evaluate its validity of stylistic dependency using a Likert scale.
Our prediction-based correlation shows 4.18 agreement on average, showing reasonable accuracy of correlations.

\begin{figure*}[h!]
\vspace{0mm}
\centering
{
\includegraphics[trim=0cm 6cm 2.5cm 0cm,clip,width=0.99\linewidth]{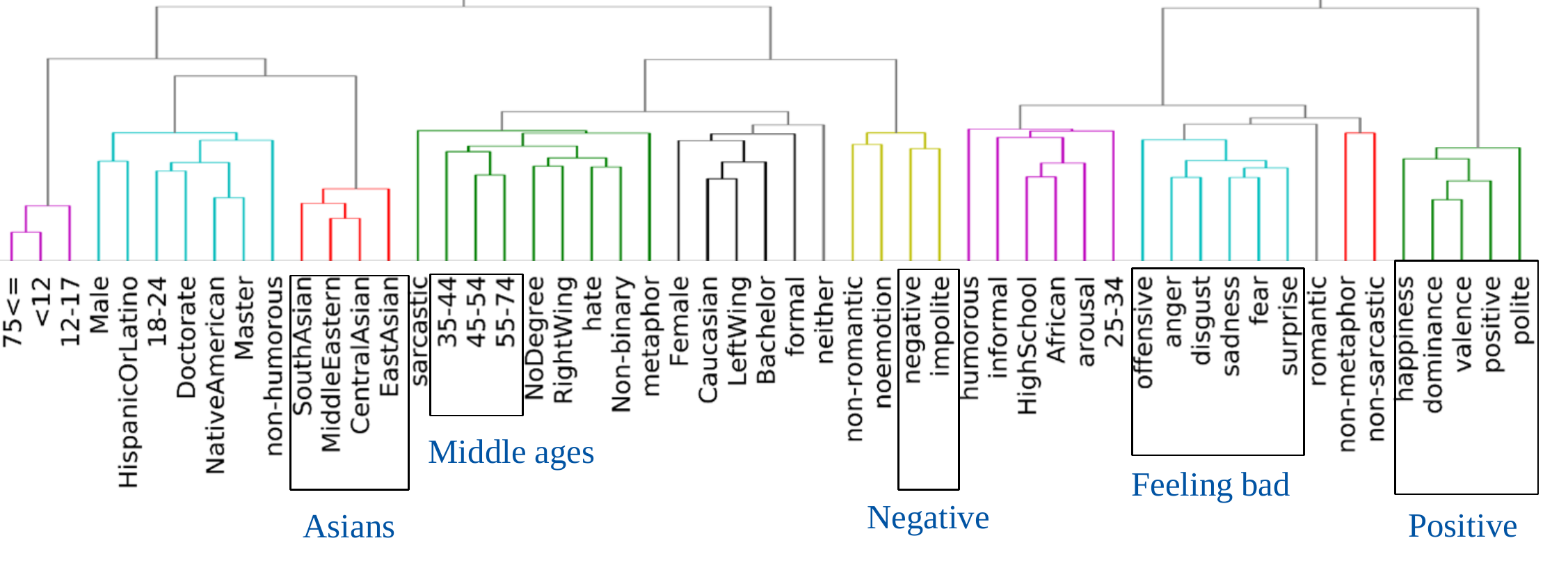}
}
\vspace{-8mm}
\caption{\label{fig:style_grouping} Empirical grouping of styles. Best viewed in color.
}
\end{figure*}

We also provide an empirical grouping of styles using Ward hierarchical clustering \cite{ward1963hierarchical} on the correlation matrix.
Figure \ref{fig:style_grouping} shows some interpretable style clusters detected from text, like Asian ethnicities (SouthAsian, EastAsian),  middle ages (35-44, 45-54, 55-74), positiveness (happiness, dominance, positive, polite), and bad emotions (anger, disgust, sadness, fear). 

\section{Case \#3: Cross-Style Generation}\label{sec:cross_style_generation}

We study the effect of combination of some styles in the context of generation.
We first describe our style-conditioned generators that combine the style classifiers in \S\ref{sec:style_classification} with pre-trained generators (\S\ref{sec:style_conditioned_generation}), and then validate two hypothetical questions using the generators: does better identification of styles help better stylistic generation (\S\ref{sec:classification_generation})? and which combination of styles are more natural or contradictive in generation (\S\ref{sec:contradictive_generation})? 

\subsection{Style-conditioned Generation}\label{sec:style_conditioned_generation}

\begin{table}[h]
\small
\begin{tabularx}{0.99\linewidth}{@{}XX@{}}
\multicolumn{2}{@{}>{\columncolor{white}[0pt][\tabcolsep]}l}{Output without style condition:}\\
\midrule
\multicolumn{2}{@{}>{\hsize=\dimexpr2\hsize+2\tabcolsep+\arrayrulewidth\relax}X@{}}{
`Every natural text is' a series of images. The images, as they are known within the text, are the primary means by which a text is read, and therefore are ..}\\
\addlinespace[0.2cm]
\multicolumn{2}{@{}>{\columncolor{white}[0pt][\tabcolsep]}l}{Output conditioned on \textbf{Formality} (F1 = 89.9\%) }\\
\multicolumn{2}{@{}>{\columncolor{white}[0pt][\tabcolsep]}l}{: {\textcolor{blue}{Formal}} (left) and {\textcolor{red}{Informal}} (right) }\\
\midrule
`Every natural text is' different. You {\textcolor{blue}{may}} find that the word you wrote does not appear on the website of the author. {\textcolor{blue}{If you have any queries}}, you can contact us.. 
& 
`Every natural text is' {\textcolor{red}{a bit}} of a hack. I don't think of it as a hack, because this hack is the hack.. and if you don't believe me then please don't read this, {\textcolor{red}{I don't care}}..
\\
\addlinespace[0.2cm]
\addlinespace[0.2cm]
\multicolumn{2}{@{}>{\columncolor{white}[0pt][\tabcolsep]}l}{Output conditioned on \textbf{Offense} (F1 = 93.4\%) }\\
\multicolumn{2}{@{}>{\columncolor{white}[0pt][\tabcolsep]}l}{: {\textcolor{blue}{Non-offensive}} (left) and {\textcolor{red}{Offensive}} (right) }\\
\midrule
`Every natural text is' a natural language, and every natural language is a language that we can speak. It is the language of our thoughts and of our lives..
&
`Every natural text is' worth reading...I'm really going to miss the music of David Byrne, and that was so much fun to watch live. The guy is a {\textcolor{red}{*ucking}} {\textcolor{red}{*ick}}. ..
\\
\end{tabularx}
\vspace{-1mm}
\caption{\label{tab:style_generation_output} Given a prompt `Every natural text is', output text predicted by our stylistic generator. The blue and red phrases are manually-labeled as reasonable features for each label. Offensive words are replaced with *.
\vspace{-1mm}} 
\end{table}

Let $x$ an input text and $s$ a target style.
Since we already have the fine-tuned style classifiers $\mathrm{P}(s|x)$ from \S\ref{sec:style_classification}, we can combine them with a generator $\mathrm{P}(x)$, like a pre-trained language model, and then generate text conditioned on the target style $\mathrm{P}(x|s)$.
We extend the plug-and-play language model (PPLM) \cite{dathathri2019plug} to combine our style classifiers trained on \method with the pre-trained generator; GPT2 \cite{radford2019language} without extra fine-tuning: $\mathrm{P}(x|s) \propto \mathrm{P}(x) \cdot \mathrm{P}(s|x)$.
Table \ref{tab:style_generation_output} shows example outputs from our style-conditioned generators given a prompt `Every natural text is'.

\begin{table}[bt]
\centering
\fontsize{8.1}{9.1}\selectfont{
\begin{tabularx}{\columnwidth}{@{}r@{\hskip 0.4cm} c @{\hskip 0.4cm}c @{\hskip 0.4cm}c @{\hskip 0.4cm}c@{}}
\toprule
&	Sentiment&	Politeness&		Formality&	Offense \\
\midrule
\method (F1) &	96.5	&71.2		&89.8	&93.3\\
\hdashline[0.4pt/2pt]
Auto (F1) &	73.7 &	70.1 &	60.0 	 &63.7 \\
\hdashline[0.4pt/2pt]
Human({\textcolor{blue}{1$^{st}$}})  & 3.4/3.5/2.8 & 3.6/3.6/3.3 & 3.4/3.7/3.1 & 4.0/3.9/3.3\\
Human({\textcolor{red}{2$^{nd}$}})  & 2.4/3.2/2.3 & 2.8/3.4/2.7 & 2.9/2.8/2.0 & 2.9/3.3/2.5\\
\bottomrule
\end{tabularx}
\vspace{-1mm}
}
\caption{\label{tab:style_gen_eval} Automatic and human evaluation on generated text. 1$^{st}$ and 2$^{nd}$ labels correspond to
{\textcolor{blue}{positive}} and {\textcolor{red}{negative}} for sentiment, {\textcolor{blue}{polite}} and {\textcolor{red}{impolite}} for politeness, {\textcolor{blue}{formal}} and {\textcolor{red}{informal}} for formality, and {\textcolor{blue}{non-offensive}} and {\textcolor{red}{offensive}} for offense. Three numbers in human evaluation means stylistic appropriateness, consistency with prompt, and overall coherence in order.\vspace{-2mm}}
\end{table}

We evaluate quality of output text:
given 20 frequent prompts randomly extracted from our training data,\footnote{Some example prompts: ``Meaning of life is'', ``I am'', ``I am looking for'', ``Humans are'', ``The virus is'', etc} we generate 10 continuation text for each prompt for each binary label of four styles (sentiment, politeness, offense, and formality)\footnote{We choose them by the two highest F1 scored styles each from inter-personal and affective groups, although we conduct experiments on other styles such as romance and emotions.} using the conditional style generator; total $20 * 10 * 2 * 4$=$1600$ continuations.

We evaluate using both automatic and human measures:
In automatic evaluation, we calculate F1 score of generated text using the fine-tuned classifiers, to check whether the output text reflects stylistic factor of the target style given.
In human evaluation, scores (1-5 Likert scale) annotated by three crowd-workers are averaged on three metrics: \textit{stylistic appropriateness}\footnote{Stylistically appropriateness means the output text includes appropriate amount of target style given.}, \textit{consistency with prompt}, and \textit{overall coherence}. 

In Table \ref{tab:style_gen_eval}, compared to F1 scores on individual test set in \method, automatic scores on output from the generator are less by 20.5\% on average, showing sub-optimality of the conditional style generator between classification and generation. 
Interestingly, in human evaluation, negative labels (2$^{nd}$ label for each style) for each style, like negative sentiment, impoliteness, informality, and offensiveness, show less stylistic appropriateness than positive or literal labels. 

\subsection{Better classification, better generation}
\label{sec:classification_generation}
\begin{figure}[t!]
\vspace{-2mm}
\centering
{
\includegraphics[trim=0cm 5.7cm 12cm  0cm,clip,width=.99\linewidth]{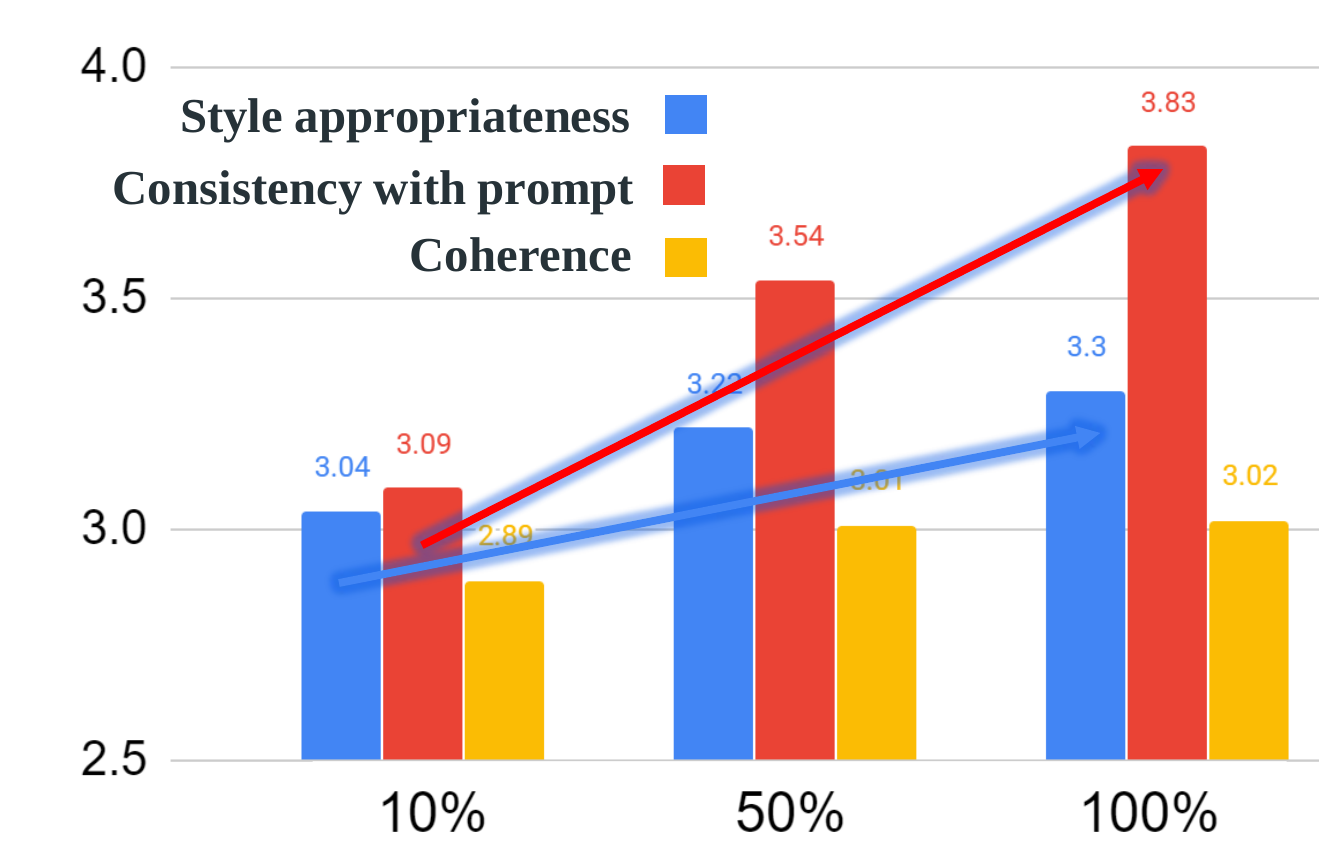}
\vspace{-1mm}
}
\caption{\label{fig:classification_vs_generation} As the training completion ratio (x-axis, \%) of classifiers increases, stylistic appropriateness (blue, y-axis) and consistency (red, y-axis) increase. }
\vspace{-1mm}
\end{figure}

To further investigate the relationship between classifier's performance and generation quality, we conduct a study by decreasing the training completion ratio (i.e., a fraction of epochs until completion; $C\%$) of the classifiers; $\mathrm{P}_{C\%}(s|x)$ over the four styles and again evaluate the output continuation; $\mathrm{P}_{C\%}(x|s) \propto \mathrm{P}(x) \cdot \mathrm{P}_{C\%}(s|x)$ using the same human metrics.
Figure \ref{fig:classification_vs_generation} shows that the better style understanding (higher F1 scores in classification) yields the better stylistic generation (higher stylistic appropriateness and consistency scores).

\subsection{Contradictive styles in generation}\label{sec:contradictive_generation}
We have generated text conditioned on single styles. 
We now generate text conditioned on combination of multiple styles; $\mathrm{P}(x|s_1..s_k) \& \propto \mathrm{P}(x) \cdot \mathrm{P}(s_1|x) \cdots \mathrm{P}(s_k|x)$
where $k$ is the number of styles.
In our experiment, we set $k$=$2$ for sentiment and politeness styles, and generate text conditioned on all possible combinations between the labels of the two styles (e.g., positive and polite label, negative and impolite label).
We again conduct human evaluation on the output text for measuring whether the generator produces stylistically appropriate text given the combination.

  \begin{table}[t!]
    \centering
\small
\begin{tabularx}{0.45\columnwidth}{@{}ccc@{}}
\toprule
& Polite & Impolite \\
\midrule
Pos & 3.11 & 2.45 \\
Neg & 2.52 & 2.89 \\
\bottomrule
\end{tabularx}
\quad
\begin{tabularx}{0.45\columnwidth}{@{}ccc@{}}
\toprule
& Polite & Impolite \\
\midrule
Pos & \cellcolor{SkyBlue!25}0.58 & \cellcolor{yellow!15}0.21 \\
Neg & \cellcolor{yellow!15}0.17 & \cellcolor{SkyBlue!25}0.63 \\
\bottomrule
\end{tabularx}
\vspace{-1mm}
    \caption{\label{tab:multi_style_genreation} Stylistic appropriateness scores (human judgement) on model-generated text with Likert scale (left) and style correlation scores from the correlation matrix (right) between politeness and sentiment.
    \vspace{-2mm}}
  \end{table}

Table \ref{tab:multi_style_genreation} shows averaged human-measured stylistic appropriate scores over the four label combinations (left) and the correlation scores observed in the style correlation matrix on written text in Figure \ref{fig:style_corr} (right).
Some combinations, like positive and impolite or like negative and polite, show less stylistic appropriateness scores, because they are naturally contradictive in their stylistic variation.
Moreover, the stylistic appropriateness scores look similar to the correlation score observed from written text, showing that there exists some natural or unnatural combination of styles in both classification on human-written text and output generated by the model.

\section{Conclusion and Discussion}
We introduce a benchmark \method of mostly existing datasets for studying cross-style language understanding and evaluation.
Using \method, we found interesting cross-style observations in classification, correlation, and generation case studies.
We believe \method helps other researchers develop more solid methods on various cross-style applications. 
We summarize other concerns we found from our case studies:

\textbf{Style drift.}
The biggest challenge in collecting style datasets is to diversify the style of text but preserve the meaning, to avoid \textit{semantic drift}. 
In the cross-style setting, we also faced a new challenge; \textit{style drift}, where different styles are coupled so changing one style might affect the others.

\textbf{Ethical consideration.}
Some styles, particularly on styles related to personal traits, are ethically sensitive, so require more careful interpretation of the results not to make any misleading points.
Any follow-up research needs to consider such ethical issues as well as provides potential weaknesses of their proposed methods.

\textbf{From correlation to causality.}
Our analysis is based on correlation, not causality. 
In order to find causal relation between styles, more sophisticated causal analyses, such as propensity score \cite{austin2011introduction}, need to be considered for controlling the confounding variables.
By doing so, we may resolve the biases driven from the specific domain of training data.
For example, generated text with the politeness classifier \cite{danescu2013computational}  contains many technical terms (e.g., 3D, OpenCV, bugs) because its training data is collected from StackExchange.

\section*{Acknowledgements} 
This work would not have been possible without the efforts of the authors who kindly share the style language datasets publicly. 
We thank Edvisees members at CMU, Hearst lab members at UC Berkeley, and anonymous reviewers for their helpful comments.


\bibliographystyle{acl_natbib}
\bibliography{xslue}

\renewcommand*\appendixpagename{\Large Appendices}
\clearpage
\begin{appendix}\label{sec:appendix_dialrec}

\section{Details on Preparation for Style Datasets in \method}\label{sec:style_dataset}

\paragraph{Formality.}
Appropriately choosing the right formality in the situation is the key aspect for effective communication \cite{heylighen1999formality}.
We use \texttt{GYAFC} dataset \cite{rao2018dear} that includes both formal and informal text collected from the web.\footnote{The dataset requires an individual authorization for access, so we only provide a script for preprocessing.}

\paragraph{Humor.}
Humor (or joke) is a social style to make the conversation more smooth or make a break \cite{rodrigosequential,kiddon2011s}. 
We use the two datasets widely used in humor detection: \texttt{ShortHumor} \cite{shorthumor}
and \texttt{ShortJoke} \cite{shortjoke}, where both are scraped from several websites. 
We randomly sample negative (i.e., non-humorous) sentences from random text from Reddit corpus \cite{kang19emnlp_biassum} and literal text from Reddit corpus \cite{khodak2017large}.

\paragraph{Politeness.}
Encoding (im)politeness in conversation often plays different roles of social interactions such as for power dynamics at workplaces, decisive factor, and strategic use of it in social context \cite{chilton1990politeness,clark1980polite}. 
We use Stanford's politeness dataset \texttt{StanfPolite} \cite{danescu2013computational} that includes request types of polite and impolite text scraped from Stack Exchange question-answer community. 

\paragraph{Sarcasm.}
Sarcasm acts by using words that mean something other than what you want to say, to insult someone, show irritation, or simply be funny.
We choose two of them for \method: \texttt{SarcGhosh} \cite{ghosh2016fracking} and \texttt{SARC} v2.0 \cite{khodak2017large}\footnote{\texttt{SARC}$_{pol}$ is a sub-task for the text from politics subreddit.}.
For \texttt{SARC}, we use the same preprocessing scheme in \citet{ilic2018deep}.

\paragraph{Metaphor.}
Metaphor is a figurative language that describes an object or an action by applying it to which is not applicable. 
We use two benchmark datasets:\footnote{we did not include  \citet{mohler2016introducing}'s dataset because the labels are not obtained from human annotators.} 
Trope Finder (\texttt{TroFi}) \cite{birke2006clustering} and VU Amsterdam \texttt{VUA} Corpus \cite{steen2010method} where metaphoric text is annotated by human annotators.

\paragraph{Offense.}
Hate speech is a speech that targets disadvantaged social groups based on group characteristics (e.g., race, gender, sexual orientation) in a manner that is potentially harmful to them \cite{jacobs1998hate,walker1994hate}.
We use the \texttt{HateOffenssive} dataset \cite{davidson2017automated} which includes hate text ($7\%$), offensive text ($76\%$), and none of them ($17\%$).

\paragraph{Romance.}
Since we could not find any dataset containing romantic texts, we crawl and preprocess them from eleven different web sites, then make a new dataset  \texttt{ShortRomance}.
We make the same number of negative samples from the literal Reddit sentences \cite{khodak2017large} as the romantic text.
\texttt{ShortRomance} text are crawled from the following websites. 
The copyright of the messages are owned by the original writer of the websites. 
\begin{itemize}[noitemsep,topsep=0pt,leftmargin=*]
\item\url{http://www.goodmorningtextmessages.com/2013/06/romantic-text-messages-for-her.html}
\item\url{https://www.travelandleisure.com/travel-tips/romantic-love-messages-for-him-and-her}
\item\url{https://www.amoramargo.com/en/sweet-text-messages-for-her/}
\item\url{https://www.techjunkie.com/best-romantic-text-messages-for-girlfriend/}
\item\url{https://liveboldandbloom.com/10/relationships/love-messages-for-wife}
\item\url{https://www.marriagefamilystrong.com/sweet-love-text-messages/}
\item\url{https://pairedlife.com/love/love-messages-for-him-and-her}
\item\url{https://truelovewords.com/sweet-love-text-messages-for-him/}
\item\url{https://www.serenataflowers.com/pollennation/love-text-messages/}
\item\url{https://www.greetingcardpoet.com/73-love-text-messages/}
\item\url{https://www.wishesmsg.com/heart-touching-love-messages/}
\end{itemize}

\paragraph{Sentiment.}
Identifying sentiment polarity of opinion is challenging because of its implicit and explicit presence in text \cite{kim2004determining,pang2008opinion}.
We use the annotated sentiment corpus on movie reviews; Sentiment Tree Bank \cite{socher2013recursive} (\texttt{SentiBank}).

\paragraph{Emotion.}
Emotion is more fine-grained modeling of sentiment.
(degree of control).
We use two datasets: \texttt{DailyDialog} \cite{li2017dailydialog} from the \citet{ekman1992argument}'s six categorical model, and \texttt{EmoBank} \cite{buechel2017emobank} from \cite{warriner2013norms}'s three-dimensional VAD model.
We also include a large but noisy emotion-annotated corpus \texttt{CrowdFlower} \cite{crowdflower}, including in addition to Ekman's categories, additional seven categories: enthusiasm, worry, love, fun, hate, relief, and boredom.

\paragraph{Persona.}
Persona is a pragmatics style in group characteristics of the speaker.
We use the stylistic language dataset written in parallel called \texttt{PASTEL} \cite{kang19emnlp_pastel} where multiple types of the author's personas are given in conjunction.
\texttt{PASTEL} has six different persona styles (i.e., age, gender, political view, ethnicity, country, education) where each has multiple attributes.



\section{Details on Annotation Schemes}\label{sec:annotation_scheme}
\begin{figure*}[ht!]
\vspace{0mm}
\centering
{
\includegraphics[trim=0cm 2cm 0.5cm  0cm,clip,width=.99\linewidth]{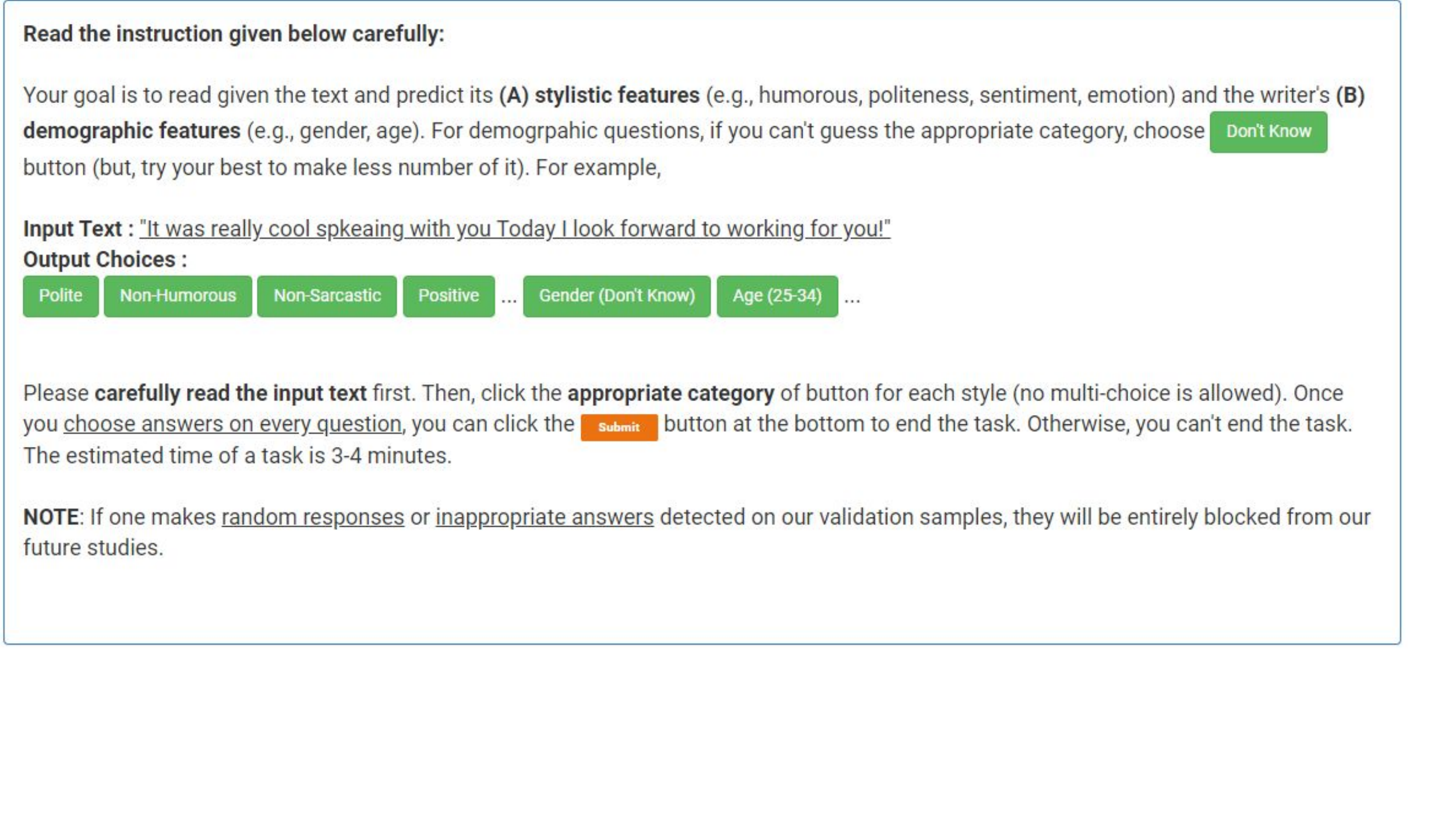}
\\
\includegraphics[trim=0cm 3cm 9cm  0cm,clip,width=.99\linewidth]{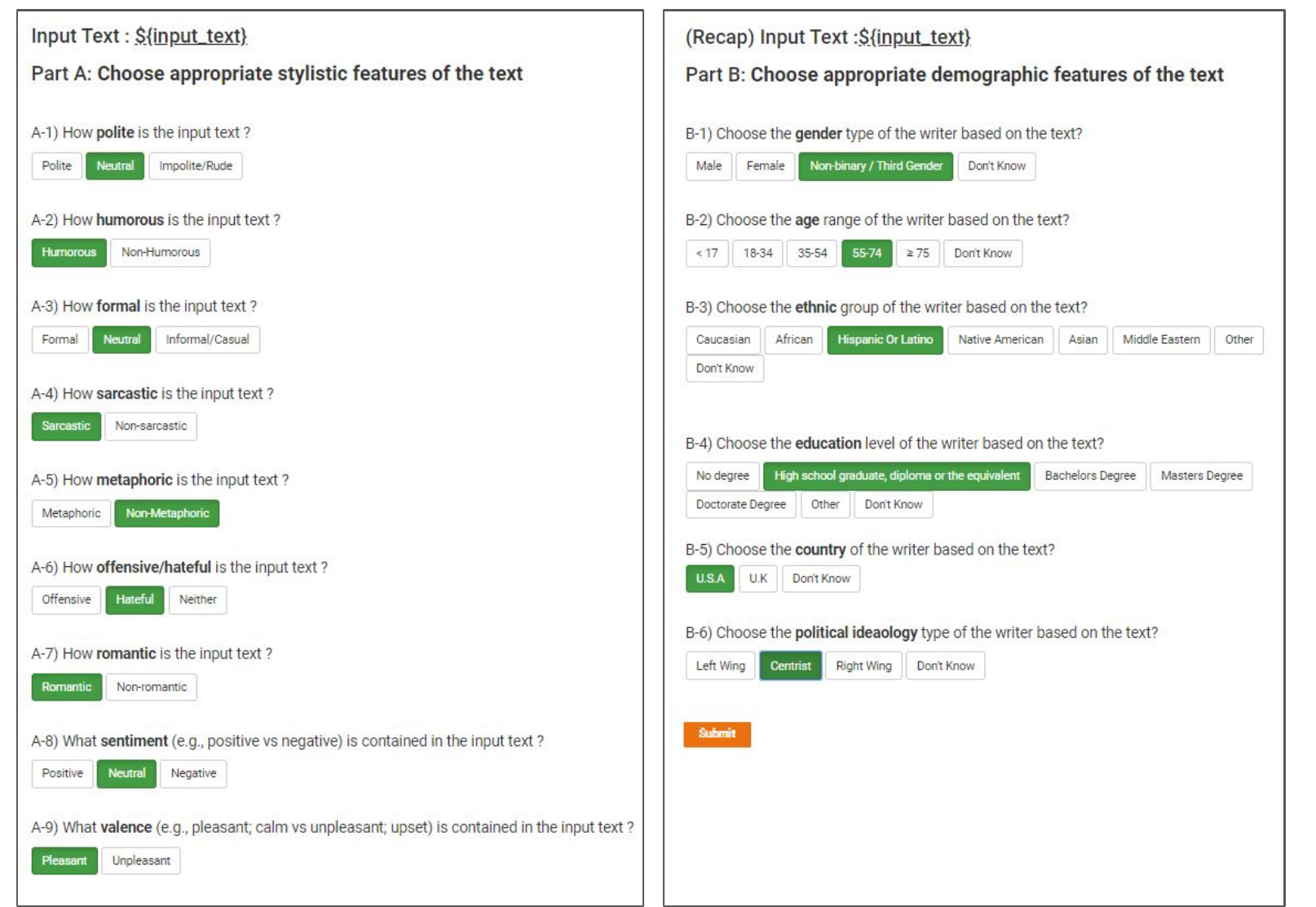}
}
\caption{\label{fig:amt_snapshot} Snapshots of our annotation tasks: general instruction (top) and annotation tasks on each style (bottom). }
\vspace{0mm}
\end{figure*}

Figure \ref{fig:amt_snapshot} shows snapshots of our annotation platform with the detailed instructions.
We estimate the execution time of a task 
as 4 minutes, so paying $\$9 / ( 60 minutes / 3 minutes) = \$0.4$ per task.
We make 10 size of batches multiple times and incrementally increase the size of batches up to 400 samples. 
For each batch, we manually checked the quality of outputs and blocked some bad users who abusively answered the questions.


\section{Details on Hyper-Parameters}\label{sec:hyper_parameters}
For our BERT classifier, we use the uncased BERT English model. 
We did parameter sweeping on learning rates [{2,5}e-{5,4}] and batch sizes [32,16,8] on the validation set. 
For the BiLSTM baseline, we use 32\footnote{32 batch size shows slightly better performance than smaller sizes like 8 or 16.} sizes of batching for both training and testing and 256 hidden size for LSTM layer with 300 sizes of word embedding from GloVe \cite{pennington2014glove}.
The vocabulary size of BiLSTM is the same as the maximum vocabulary of the BERT model; 30522. 
For both BERT and BiLSTM models, we use the same maximum input length 128. 
Both training use $2e-5$ learning rate and $1.0$ maximum gradient clipping with Adam optimizer with $1e-8$ epsilon. 
Also, we use early stopping until the maximum training epochs of 5.

\end{appendix}

\end{document}